\documentclass{article}

    \usepackage[preprint]{neurips_2023}

\usepackage{natbib}
\setcitestyle{numbers,square}

\usepackage[utf8]{inputenc} 
\usepackage[T1]{fontenc}    
\usepackage{hyperref}       
\usepackage{url}            
\usepackage{booktabs}       
\usepackage{amsfonts}       
\usepackage{appendix}       
\usepackage{tikz}           

\usepackage{nicefrac}       
\usepackage{microtype}      
\usepackage{xcolor}         
\usepackage{graphicx}       

\title{Machine Mindset: An MBTI Exploration of Large Language Models}

\author{
    Jiaxi Cui\thanks{Equal Contribution.}  \\
    Panda Villa Tech   \\
    Peking University   \\
    \texttt{jiaxicui446@gmail.com}   \\
    \And
    Liuzhenghao Lv$^*$   \\
    Peking University   \\
    \texttt{lvliuzh@stu.pku.edu.cn} \\
    \And
    Jing Wen   \\
    Panda Villa Tech   \\
    \texttt{wenjing98vv@163.com} \\
    \And
    Rongsheng Wang   \\
    Macao Polytechnic University   \\
    \texttt{p2213046@mpu.edu.mo} \\
    \And
    Jing Tang   \\
    Panda Villa Tech   \\
    Huazhong University of Science and Technology   \\
    \texttt{j\_tang@hust.edu.cn} \\
    \And
    YongHong Tian  \\
    Peking University   \\
    PengCheng Laboratory, China   \\
    \texttt{yhtian@pku.edu.cn}   \\ 
    \And
    Li Yuan\thanks{Corresponding Author}  \\
    Peking University   \\
    \texttt{yuanli-ece@pku.edu.cn}   \\ 
}

\begin{document}

\maketitle

\begin{figure}[htp]
 \centering
 \includegraphics[width=1.0\columnwidth]{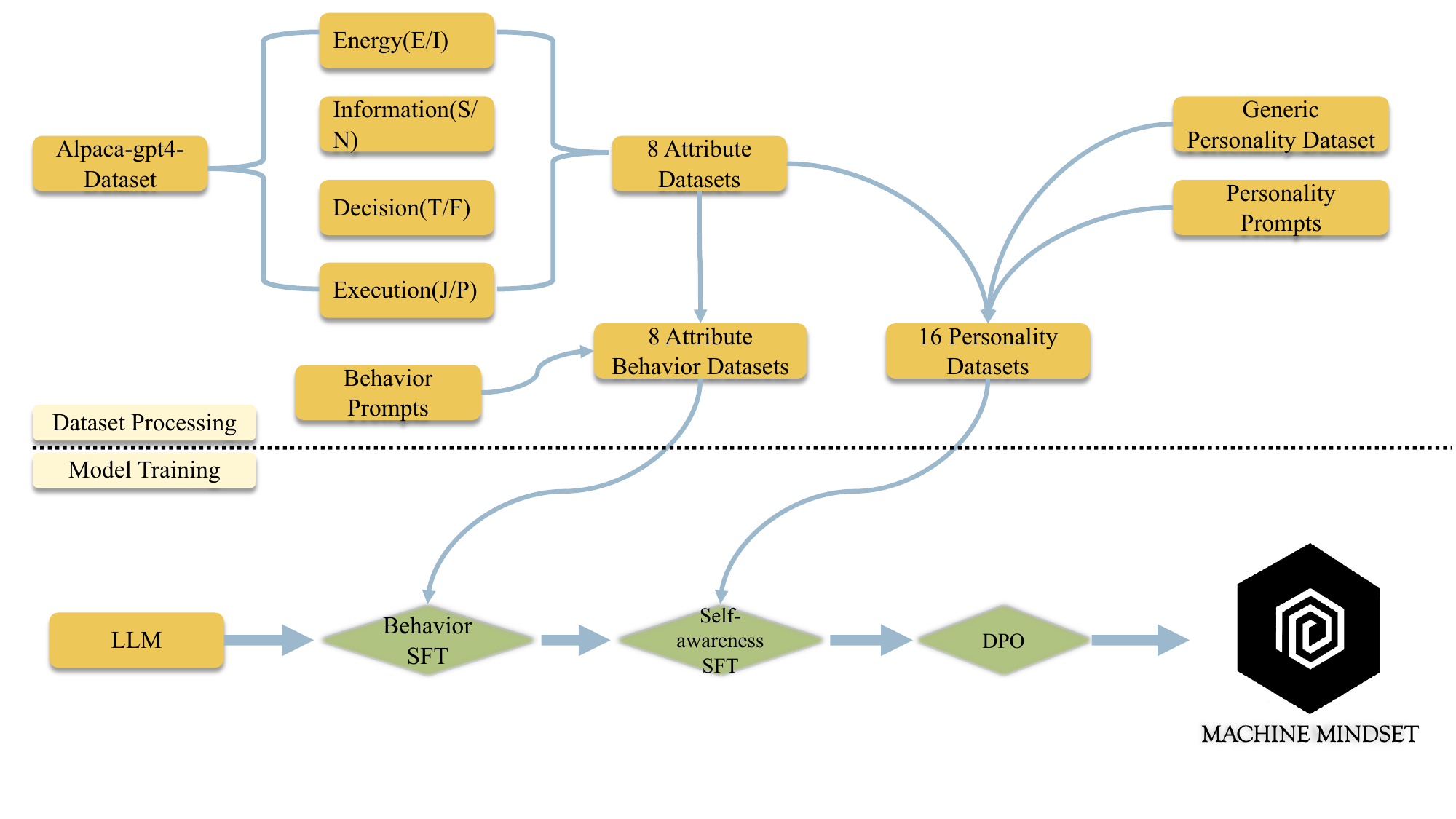}
 \caption{Machine Mindset Framework}
 \label{fig:motivation}
\end{figure}

\begin{abstract}
We present a novel approach for integrating Myers-Briggs Type Indicator (MBTI) personality traits into large language models (LLMs), addressing the challenges of personality consistency in personalized AI. Our method, "Machine Mindset," involves a two-phase fine-tuning and Direct Preference Optimization (DPO) to embed MBTI traits into LLMs. This approach ensures that models internalize these traits, offering a stable and consistent personality profile. We demonstrate the effectiveness of our models across various domains, showing alignment between model performance and their respective MBTI traits. The paper highlights significant contributions in the development of personality datasets and a new training methodology for personality integration in LLMs, enhancing the potential for personalized AI applications. We also open-sourced our model and part of the data at \url{https://github.com/PKU-YuanGroup/Machine-Mindset}.
\end{abstract}

\section{Introduction}
The emergence of models like ChatGPT has garnered widespread attention to the potential and applications of large language models (LLMs). In this context, the development of domain-specific LLMs, such as ChatLaw~\cite{cui2023chatlaw} for legal applications and BloombergGPT~\cite{wu2023bloomberggpt}, for the financial sector, has also become a prominent research focus. Beyond constructing foundational generic models, researchers have shifted their focus towards domain-specific LLMs, models capable of handling long-form text, and the realm of personalized models~\cite{zhao2023survey}. Additionally, customization of models in the style of the Myers-Briggs Type Indicator (MBTI) has emerged as a compelling research direction~\cite{rao2023can}.

However, prior work has been constrained by the lack of corresponding datasets, often resorting to direct MBTI tests on LLMs or attempting to induce different personality traits through simplistic prompts. Unfortunately, these methods frequently fail to achieve stable personality adjustments, necessitating the need for a more rational and effective approach.

In this work, we introduce an innovative approach to tackle the challenges associated with data construction and model training in the context of personalized models. We focus on the concept of "Machine Mindset." The core idea revolves around injecting specific MBTI personality types into the model through a two-phase fine-tuning followed by a single-stage Direct Preference Optimization (DPO). Unlike mere prompt modifications, our trained models internalize personality traits, avoiding issues like personality disarray. Furthermore, the model's capabilities vary depending on the specific MBTI personality type, which is similar to how human abilities can be influenced by personality.

In the subsequent sections of this paper, we will provide detailed explanations of our data construction methods, model training specifics, and present relevant experimental results. We will also make comparisons with traditional approaches and explore potential application domains and future research directions.

Our research methodology has undergone extensive testing across various domains, including law, patents, general aptitude tests, IQ assessments, and more. Experimental results demonstrate that the performance of different personality models aligns well with their corresponding personality traits. Moreover, our training data and processes remain consistent, reducing reliance on specific LLMs and facilitating the convenient integration of our approach with new, updated, or general-purpose LLMs. Consequently, our research offers an innovative and efficient solution to the challenges of personalization in LLMs, while simultaneously reducing the cost of interpersonal understanding and bridging social distances.

The primary contributions of this paper can be summarized as follows:
\begin{itemize}
    \item We propose a method for constructing personality datasets. Based on this, we build behavior and self-awareness datasets, respectively.
    \item We propose a training method for injecting a specific personality into the model. This includes a two-stage supervised fine-tuning and direct preference optimization.
    \item The model we train is capable of learning specific behavioral patterns associated with a particular personality and acquiring self-awareness corresponding to that personality.
\end{itemize}

\section{Related Work}
Previous studies have engaged in testing the personalities of large language models, primarily through prompting techniques ~\cite{huang2023ceval,hendrycks2021measuring}. These attempts usually rely on standard personality testing methods adapted for language models. However, the stability and clarity of the derived personality traits have remained topics of contention.

\subsection{MBTI: Myers-Briggs Type Indicator}
The Myers-Briggs Type Indicator (MBTI) is a widely recognized and influential psychological assessment tool used to categorize individuals into specific personality types. Developed by Katharine Cook Briggs and her daughter Isabel Briggs Myers ~\cite{boyle1995myers}, the MBTI is based on Carl Jung's theory of personality. It classifies individuals into one of sixteen personality types, each characterized by four dichotomous dimensions:

\begin{itemize}
  \item Energy: Extraversion (E)——Introversion (I)
  \item Information: Sensing (S)——Intuition (N)
  \item Decision: Thinking (T)——Feeling (F)
  \item Execution: Judging (J)——Perceiving (P)
\end{itemize}

The MBTI provides insights into how individuals perceive and interact with the world, make decisions, and process information. Each personality type is a unique combination of these four dimensions, resulting in diverse personality profiles. These personality types have been widely used in various fields, including psychology, career counseling, and personal development.

\subsection{General LLMs}
In the domain of general LLMs, prior research has examined the diversity of personality types exhibited by different models. One notable study ~\cite{pan2023llms} tasked various LLMs with completing MBTI assessment questions, aiming to assess and compare the personalities manifested by these models. This work represents an initial step towards understanding how different LLM architectures and training data can influence the expression of personality traits within the context of MBTI.



\section{Method}

\subsection{Dataset Construction}
\begin{figure}[ht]
  \centering
  \includegraphics[width=440pt, height=220pt]{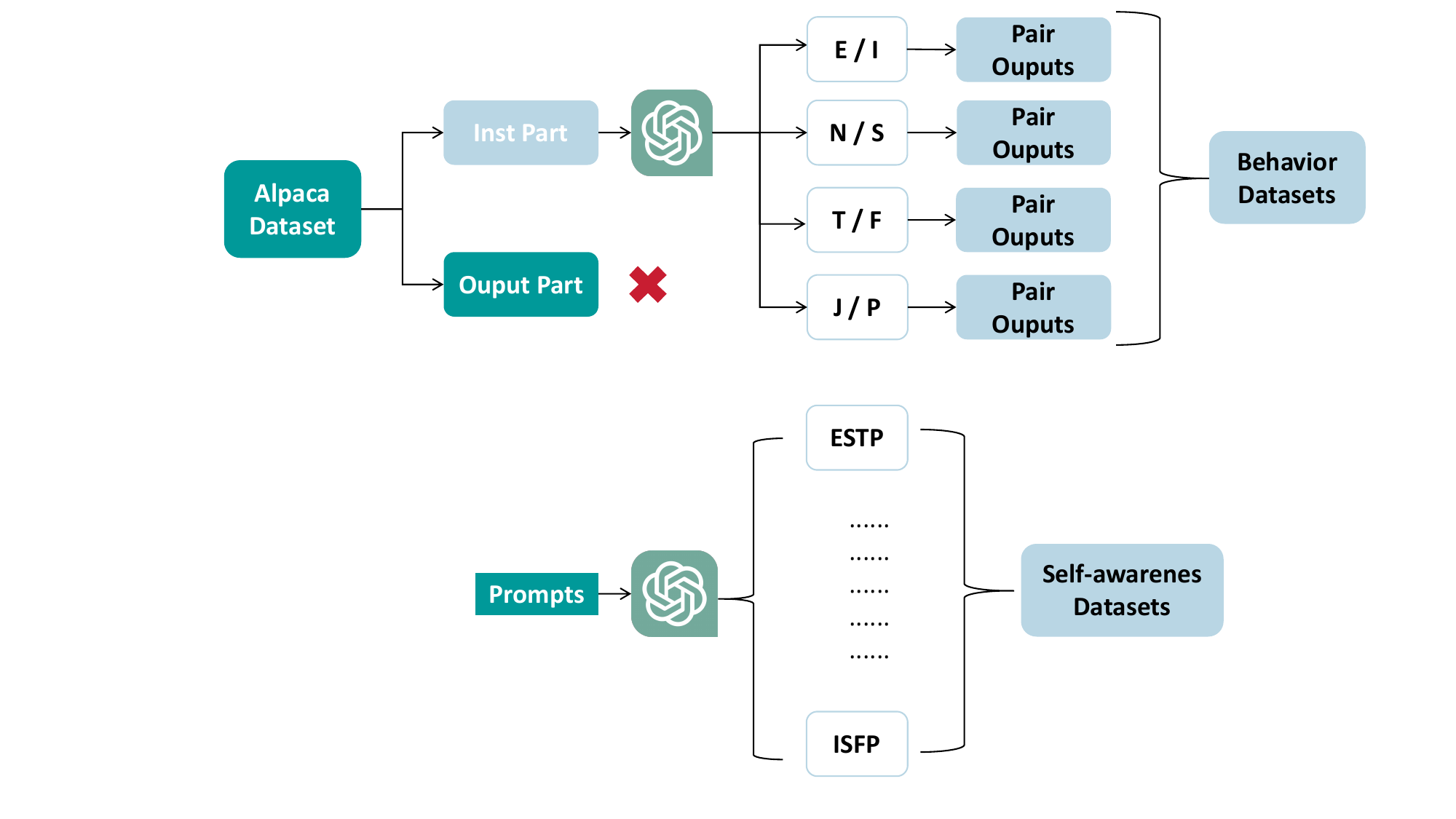}
  \caption{The flowchart for constructing behavior datasets and Self-awareness datasets.}
  \label{data}
\end{figure}

We have constructed two types of datasets: \textbf{behavior datasets} and \textbf{self-awareness datasets}. The purpose of behavior datasets is to train LLMs to respond to general user instructions consistently with specific personality traits. To ensure broad coverage across various domains, we chose to customize the Alpaca dataset \cite{alpaca,peng2023gpt4llm} for personality-specific modifications.

On the other hand, self-awareness datasets aim to enable LLMs to accurately recognize their own personality traits. This need arises from the observation that humans sometimes struggle to accurately summarize their own personality traits, indicating vague self-awareness. This phenomenon is even more pronounced in children. Similarly, we hypothesize that LLMs trained solely on behavior datasets, while capable of generating responses reflecting personality traits, lack precise self-awareness of these traits. Hence, the construction of self-awareness datasets is essential. The flowchart of building the two types of datasets is shown in Figure~\ref{data}. Details are shown below.

\begin{figure}[ht]
  \centering
  \includegraphics[width=200pt, height=120pt]{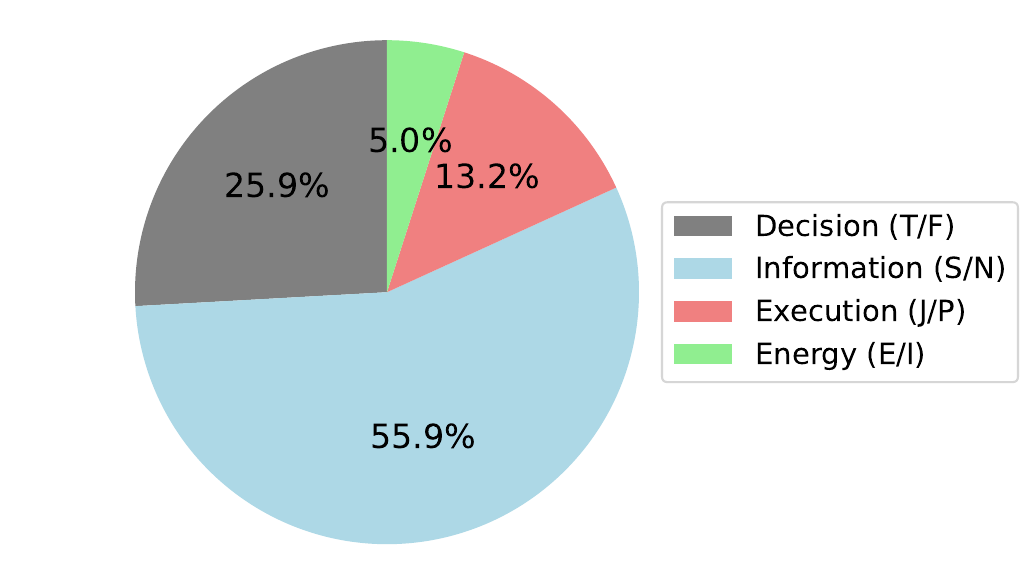}
  \caption{The ratio of the four MBTI dimensions in behavior datasets}
  \label{ratio}
\end{figure}
\textbf{Behavior Datasets.}
We meticulously processed the Alpaca dataset to create a dataset tailored for fine-tuning LLMs to exhibit different personalities. For each instruction in each Alpaca data entry, we engaged ChatGPT in a classification task to determine the most fitting MBTI dimension out of the four. Subsequently, ChatGPT generated a pair of responses to the same instruction, each mirroring one of the two attitudes within the identified dimension. Following this process, we obtained diverse Alpaca datasets conducive to the supervised fine-tuning of LLMs, enabling them to emulate distinct personalities. The composition ratios of these datasets, as depicted in Figure \ref{ratio}, reveal that the "Energy" dimension is minimally represented, while the "Information" dimension is predominant. This distribution suggests that the "Information" dimension exerts a significantly greater influence, whereas the "Energy" dimension has a lesser impact.
We refer to the datasets resulting from this process as behavior datasets, given their purpose to enable LLMs to generate language responses corresponding to different personality traits when responding to user instructions.

\textbf{Self-awareness Datasets.}
To bridge the gap between the behavior and self-awareness, we build self-awareness datasets comprising a series of Q\&As tailored to elucidate the characteristics of sixteen personality types of MBTI.
The majority of questions in these Q\&As are direct or indirect inquiries about personalities, while the answers involve descriptions of one's own personalities. These datasets are generated ChatGPT guided by certain prompts.

\subsection{Fine-tuning LLMs towards a certain personality}
We imparted different personalities to LLMs through supervised fine-tuning, employing a two-stage supervised fine-tuning process using both the behavior and self-awareness datasets, respectively. As an example, for the "INFP" personality type, we utilized four datasets corresponding to "I," "N," "F," and "P" from the behavior datasets for the first stage of supervised fine-tuning. Additionally, we retrieved an extra dataset aimed at enhancing self-awareness as an INFP individual from the self-awareness datasets for the second stage of supervised fine-tuning. Ultimately, the trained LLM exhibited behavior and self-awareness similar to those of an INFP individual.

We employed Low Rank Adaptation (LoRA) for its efficiency in supervised fine-tuning. Unlike full-parameter fine-tuning, LoRA is computationally economical and modular, allowing for the creation of personality-specific plug-ins for an LLM. This modularity facilitates seamless transitions between personality types, such as from INFP to INTJ, by merely exchanging the corresponding LoRA adapters."

\textbf{Direct Preference Optimization.}
DPO is intended to serve as an alternative to Reinforcement Learning from Human Feedback (RLHF) in LLM alignment. DPO facilitates alignment by enabling the LLM to distinguish a preferred response from a given pair. We believe that DPO is well-suited for LLMs to learn the distinctions between a pair of attitudes in each dimension of MBTI. The process involves obtaining datasets of the two attitudes within one dimension from our datasets, such as "F" and "T" in the "Decision" dimension. Subsequently, DPO is conducted, allowing the LLM to prefer "F" responses over "T" responses.

\subsection{Evaluation Method}
We evaluated the trained LLMs using the existing MBTI questionnaire with slight modifications. These modifications were made to enhance the clarity and comprehensibility of the question descriptions without altering their original intent. The rationale behind these changes stems from the observation that some of the original question descriptions were not very clear, and LLMs, especially those with fewer parameters like 7B models, sometimes struggled to grasp the intended meaning. Since the MBTI questionnaire aims to assess the personality traits of LLMs, ensuring that understanding the questions is not an obstacle is crucial.

The primary objective of the questionnaire test mentioned above was to demonstrate that our trained model exhibits the desired personality traits. However, it is essential to note that the results of this test should not be considered definitive due to the limitations of the questionnaire, including its multiple-choice format. These results should be regarded as a reference point.

Additionally, we conducted supplementary tests to investigate how various personality traits impact the performance of LLMs. Detailed information on these tests can be found in the experimental section.




\section{Experiments and Results\label{sec:experiments}}
\subsection{Training Process}
In our research, we embarked on a comprehensive training process to infuse distinct personality traits into large language models (LLMs). This process was carried out through a combination of Supervised Fine-Tuning (SFT) and Direct Preference Optimization (DPO) using our carefully curated dataset. The primary aim was to endow these models with unique and stable personalities.

During the training, we meticulously fine-tuned LLMs for each of the sixteen MBTI personality types, encompassing both English and Chinese language variants. The two-stage SFT process refined their language generation capabilities, while the subsequent DPO phase reinforced specific personality traits, fostering a deeper integration of these traits into the models.
\subsection{Evaluation Results}
In this section, we present the evaluation results of LLMs trained on our dataset. The evaluation process encompasses a series of tests designed to measure the efficacy and authenticity of the imbued personality traits.

Our evaluation criteria include:

\textbf{MBTI Questionnaire Testing}
To assess if the LLMs exhibit personalities consistent with their assigned MBTI types. These tests employ slight modifications to the traditional MBTI questionnaire to account for potential comprehension issues.

\textbf{Performance Metrics}
We applied performance metrics to measure the language generation quality, coherence, and relevance of responses generated by the personality-tuned models across various tasks and domains.

\textbf{User Feedback Assessment}
We collected user feedback through surveys and qualitative assessments to gauge user perceptions and satisfaction when interacting with different personality-tuned LLMs.

\subsection{Ablation}
To gain deeper insights into the impact of dataset compositions on the training process, we conducted ablation experiments. Specifically, we explored how the balance or imbalance in training data for the eight MBTI personality dimensions influenced the resulting personality-tuned models. By systematically varying the dataset compositions, we aimed to uncover any correlations between data distribution and the models' personality traits, as well as their performance in different scenarios.
\subsection{Effects on Ablities}
An intriguing aspect of our research focused on investigating whether the imbued personality traits had any discernible effects on the abilities of the LLMs, such as reasoning and understanding. By subjecting the personality-tuned models to a battery of tasks and tests, we sought to elucidate whether specific personality traits influenced their cognitive capabilities. This exploration provides valuable insights into the interplay between personality and cognitive functions in LLMs.

In the subsequent sections, we delve into the specific details and outcomes of each evaluation and ablation experiment, shedding light on the effectiveness and implications of our approach in imbuing LLMs with distinct and stable personality traits.

\section{Conclusion}
In this paper, we have explored the intriguing intersection of large language models (LLMs) and the Myers-Briggs Type Indicator (MBTI), aiming to imbue these powerful models with distinct and stable personality traits. Through a novel approach involving Supervised Fine-Tuning (SFT) and Direct Preference Optimization (DPO), we have successfully cultivated a diverse range of LLMs, each representing one of the sixteen MBTI personality types in both English and Chinese.

Our experiments and evaluations have yielded promising results. The LLMs trained using our methodology have exhibited personalities that align with their designated MBTI types. This achievement not only contributes to advancing the field of AI personalization but also opens up new avenues for applications in various domains, including natural language understanding, dialogue generation, and human-computer interaction.

The ablation experiments further illuminate the significance of dataset compositions in shaping the personality-tuned models. By examining the effects of data distribution on personality traits and performance, we provide valuable insights into the training dynamics of LLMs and their adaptability to diverse personality profiles.

Furthermore, our exploration into the impact of personality traits on the cognitive abilities of LLMs underscores the intricate relationship between personality and reasoning, shedding light on potential areas for future research.

It is essential to acknowledge the limitations of our study. While we have made significant strides in imbuing LLMs with personality, our approach is not without challenges. The use of the MBTI questionnaire, with its inherent limitations, leaves room for further refinement in assessing personality traits. Additionally, our experiments have primarily focused on text-based tasks, and extending this research to multimodal applications could yield intriguing results.

In conclusion, our research represents a promising step towards enhancing the personalization and adaptability of LLMs. The ability to tailor these models to exhibit distinct personalities holds immense potential for creating more engaging and relatable AI systems. We hope that our findings will inspire further exploration in the realm of personalized AI and contribute to the development of more human-like and context-aware language models.

\bibliographystyle{plain}
\bibliography{ref}

\newpage

\begin{appendices}

\section{Details on Experiments}\label{sec:appA}
\subsection{MBTI Testing Results}
We present the results obtained after two-stage self-awareness training, where each model underwent MBTI testing. The figures below showcase the performance of the models in responding to MBTI assessment questions, reflecting their acquired personality traits.

\foreach \type in {INTJ, INTP, ENTJ, ENTP, INFJ, INFP, ENFJ, ENFP, ISTJ, ISFJ, ESTJ, ESFJ, ISTP, ISFP, ESTP, ESFP}{
  \begin{figure}[htp]
    \centering
    \begin{minipage}[b]{0.65\textwidth} 
      \includegraphics[width=1.0\linewidth]{res_images/res_\type}
      \caption{MBTI Personality Type Results for \type.}
      \label{fig:mbti_\type}
    \end{minipage}
    \hfill 
    \begin{minipage}[b]{0.3\textwidth} 
      \includegraphics[width=1.0\linewidth]{mbti_res_images/\type} 
      \caption{Second Results for \type.}
      \label{fig:second_\type}
    \end{minipage}
  \end{figure}
}

\newpage
\subsection{Random Question-Answering Results}
In addition to MBTI testing, we also evaluated the models in a random question-answering task. This task involved asking each model various questions on different topics to assess their general knowledge and response quality.

\foreach \model in {INFP, ENFP, ENTJ, ESTJ, INFJ, INTP}{
  \begin{figure}[htp]
    \centering
    \includegraphics[width=1.0\columnwidth]{random_qa/\model_res}
    \caption{Random Question-Answering Results for \model.}
    \label{fig:random_qa_\model}
  \end{figure}
}

The figures above display the performance of each model in responding to a diverse set of questions. These results provide insights into how well the models can handle general inquiries and engage in open-ended conversations.

\end{appendices}
\end{document}